\newcommand{\CLASSINPUTtoptextmargin}{2.4cm}
\newcommand{\CLASSINPUTbottomtextmargin}{4.6cm}
\documentclass[conference, 10 pt, letterpaper]{IEEEtran}  

\IEEEoverridecommandlockouts 

\usepackage[english]{babel}
\usepackage[utf8]{inputenc}
\usepackage{times} 
\usepackage{cite}
\usepackage{comment}
\usepackage{balance}

\usepackage{hyperref}

\usepackage[nolist]{acronym}

\usepackage[singlespacing]{setspace} 

\usepackage{multirow}

\usepackage{siunitx}
\usepackage{graphicx} 
\usepackage{epsfig} 
\usepackage{pgfplots}
\usepackage{caption}
\usepackage[font=footnotesize]{subcaption}
\usepackage{pgfplots}

\usepackage{float}

\usepackage[export]{adjustbox}

\usepackage{mathtools}
\usepackage{nicefrac}
\usepackage{amsmath}
\usepackage{amssymb}
\usepackage{bm} 
\usepackage{relsize} 

\DeclareMathOperator*{\maximize}{maximize}

\usepackage{etoolbox}
\makeatletter%
\AfterPreamble{%
	\usepackage{hyperref}%
	\let\oldhypertarget\hypertarget%
	\renewcommand{\hypertarget}[2]{%
		\oldhypertarget{#1}{#2}%
		\protected@write\@mainaux{}{%
			\string\expandafter\string\gdef%
			\string\csname\string\detokenize{#1}\string\endcsname{#2}%
		}%
	}%
	\newcommand{\myhyperlink}[1]{%
		\hyperlink{#1}{\csname #1\endcsname}%
	}%
}
\makeatother%

\newcounter{Remark}
\newcounter{Problem}
\newcounter{Criterion}
\newcommand{\displayCriterion}[2][]{%
	\stepcounter{Criterion}%
	\textit{Criterion~}\hypertarget{#1}{\theCriterion}\textit{~(#2)}%
}

\newcommand{\refCriterion}[1][]{%
	Criterion~\myhyperlink{#1}%
}

\usepackage{algorithm}
\usepackage[noend]{algpseudocode}

\makeatletter
\def\BState{\State\hskip-\ALG@thistlm}
\makeatother

\usepackage{tikz}
\pgfdeclarelayer{foreground}
\pgfsetlayers{background, main, foreground}
\usetikzlibrary{quotes, angles, backgrounds, arrows, automata, shapes, positioning, calc, through, spy, decorations.markings}

\usepackage{aircraftshapes} 

\tikzset{
	imglabel/.style={
		rectangle,
		inner sep=2pt,
		text=black,
		minimum height=1em,
		text centered,
		fill=white,
		fill opacity=1.0,
		text opacity=1,
		anchor=south west,
	},
}

\tikzset{
	state/.style={
		rectangle,
		draw=black, very thick,
		minimum height=1.0em,
		text centered,
	},
}

\graphicspath{{./figures/}}

\newcommand\copyrighttext{%
    \small \begin{center} \color{red} \textcopyright\,2024 IEEE. Accepted for presentation to the ``2024 IEEE Global Communications Conference (IEEE GLOBECOM)", Cape Town, South Africa. Personal use of this material is permitted. Permission from IEEE must be obtained for all other uses, in any current or future media, including reprinting/republishing this material for advertising or promotional purposes, creating new collective works, for resale or redistribution to servers or lists, or reuse of any copyrighted component of this work in other works. \end{center}}
\newcommand\copyrightnotice{%
	\begin{tikzpicture}[remember picture,overlay]
	\node[anchor=south,yshift=25.6cm] at (current page.south) 
	{\color{red}\fbox{\parbox{\dimexpr\textwidth-\fboxsep-\fboxrule\relax}{\copyrighttext}}};
	\end{tikzpicture}%
}

\title{\copyrightnotice \Large \bf Harnessing the Potential of Omnidirectional Multi-Rotor Aerial Vehicles in Cooperative Jamming Against Eavesdropping
}

\author{Daniel Bonilla Licea$^{1,2}$, Hajar El Hammouti$^{1}$, Giuseppe Silano$^{2}$, and Martin Saska$^{2}$
    \thanks{$^1$Daniel Bonilla Licea and Hajar El Hammouti are with the College of Computing, Mohammed VI Polytechnic University, Ben Guerir, Morocco (emails: {\tt\small \{daniel.bonilla, hajar.elhammouti\}@um6p.ma}).}
    \thanks{$^2$Daniel Bonilla Licea, Giuseppe Silano, and Martin Saska are with the Czech Technical University in Prague, Czech Republic (emails: {\tt\small \{bonildan, silangiu, martin.saska\}@fel.cvut.cz}).}
    \thanks{This work was partially funded by the European Union under the project Robotics and Advanced Industrial Production (reg. no. CZ.02.01.01/00/22\_008/0004590), by the Czech Science Foundation (GAČR) under research project no. 23-07517S, and by CTU grant no SGS23/177/OHK3/3T/13.}
}  

\begin{document}

\maketitle
\thispagestyle{empty}
\pagestyle{empty}


\begin{acronym}
    \acro{AoA}[AoA]{Angle of Arrival}
    \acro{AoD}[AoD]{Angle of Departure}
    \acro{BS}[BS]{Base Station}
    \acro{DoF}[DoF]{Degree of Freedom}
    \acro{FDMA}[FDMA]{Frequency-Division Multiple Access}
    \acro{LoS}[LoS]{Line of Sight}
    \acro{MRAV}[MRAV]{Multi-Rotor Aerial Vehicle}
    \acro{PDF}[PDF]{Probability Density Function}
    \acro{UAV}[UAV]{Unmanned Aerial Vehicle} 
    \acro{SNR}[SNR]{Signal-to-Noise Ratio}
    \acro{SINR}[SINR]{Signal-to-Interference-plus-Noise Ratio}
    \acro{TDMA}[TDMA]{Temporal Division Multiple Access}
    \acro{wrt}[w.r.t.]{with respect to}
\end{acronym}



\begin{abstract}
  
  Recent research in communications-aware robotics has been propelled by advancements in 5G and emerging 6G technologies. This field now includes the integration of \acfp{MRAV} into cellular networks, with a specific focus on under-actuated \acsp{MRAV}. These vehicles face challenges in independently controlling position and orientation due to their limited control inputs, 
  which adversely affects communication metrics such as \acl{SNR}. In response, a newer class of omnidirectional \acsp{MRAV} has been developed, which can control both position and orientation simultaneously by tilting their propellers. However, exploiting this capability fully requires sophisticated motion planning techniques. This paper presents a novel application of omnidirectional \acsp{MRAV} designed to enhance communication security and thwart eavesdropping. It proposes a strategy where one \acs{MRAV} functions as an aerial \acl{BS}, while another acts as a friendly jammer to secure communications. This study is the first to apply such a strategy to \acsp{MRAV} in scenarios involving eavesdroppers.
  
\end{abstract}



\begin{IEEEkeywords}
    UAVs, multi-rotor systems, communication-aware robotics, jamming, eavesdropper, physical layer security
\end{IEEEkeywords}



\section{Introduction}
\label{sec:introduction}

Recent interest in communications-aware robotics has surged, evidenced by an increasing number of publications \cite{Licea2020TRO,  Wu2018TWC}.
This promising field has largely been propelled by advancements in 5G and the development of 6G technologies. These innovations aim to integrate \acfp{MRAV}  into cellular communications networks to boost performance \cite{BonillaPIEEE2024, Muralidharan2021ARCRAS, Licea2021EUSIPCO}. 

A significant portion of this research focuses on under-actuated \acp{MRAV} \cite{Jung2010CM,hammouti2018air, CalvoFullana2021IEEECM}. Such \acp{MRAV} \cite{Hamandi2021IJRR}, which can hover at specific positions and track trajectories, are pivotal for functions like mobile communications relays or aerial \acfp{BS} \cite{Licea2023ICUAS, Kishk2020VTM}. However, their primary limitation lies in the inability to independently control both position and orientation. This constraint arises from the fact that under-actuated \acp{MRAV} typically have fewer control inputs, which include thrust, roll and pitch angles, and yaw rate \cite{NascimentoARC2019}, than the number of \acp{DoF} needed for fully autonomous position and orientation control \cite{Hamandi2021IJRR}. This restricts the optimization of aerial communications networks as effective signal reception depends on both the \ac{MRAV}’s position and its orientation.

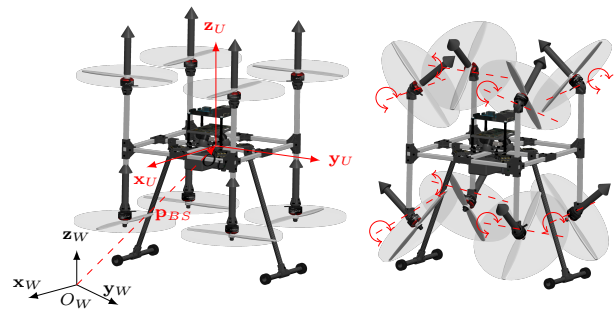
\begin{figure}[tb]
    \centering
    \scalebox{0.925}{
    \begin{tikzpicture}
        \node at (0,0) [text centered] {\adjincludegraphics[trim={{.0\width} {.0\height} {.0\width} {.0\height}}, clip, width=0.9\columnwidth]{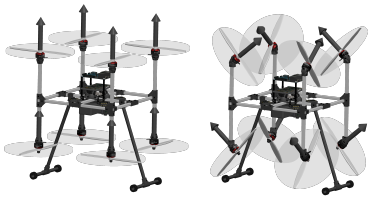}};

        \draw[-latex,red] (-1.85,0) -- (-1.85,1.5) node[above]{{\scriptsize{$\mathbf{z}_U$}}}; 
        \draw[-latex,red] (-1.85,0) -- (-0.35,-0.21) node[right]{{\scriptsize{$\mathbf{y}_U$}}}; 
        \draw[-latex,red] (-1.85,0) -- (-2.85,-0.31) node[below,red]{{\scriptsize{$\mathbf{x}_U$}}}; 
        \node at (-1.85,0) [below]{{\scriptsize{$O_U$}}};

        \draw[-latex] (-3.85,-2) -- (-3.85,-1.5) node[above]{{\scriptsize{$\mathbf{z}_W$}}}; 
        \draw[-latex] (-3.85,-2) -- (-3.25,-2.29) node[above]{{\scriptsize{$\mathbf{y}_W$}}}; 
        \draw[-latex] (-3.85,-2) -- (-4.55,-2.19) node[above]{{\scriptsize{$\mathbf{x}_W$}}}; 
        \node at (-3.85,-2) [below]{{\scriptsize{$O_W$}}};

        \draw[dashed, red] (0.60,0.66) -- (1.45,0.96); 
        \draw[<->, red]  (0.75,0.69) arc[start angle=0,end angle=280,radius=0.15];
        \draw[dashed, red] (1.2,1.16) -- (2.45,0.96); 
        \draw[<->, red]  (1.25,1.28) arc[start angle=90,end angle=-70,radius=0.15];
        \draw[dashed, red] (2.00,0.81) -- (3.15,0.31); 
        \draw[<->, red]  (2.20,0.78) arc[start angle=0,end angle=280,radius=0.15];
        \draw[dashed, red] (2.90,0.66) -- (3.70,1.01); 
        \draw[<->, red]  (3.80,0.96) arc[start angle=0,end angle=280,radius=0.15];
        \draw[dashed, red] (0.45,-1.26) -- (1.45,-0.66); 
        \draw[<->, red]  (0.66,-1.23) arc[start angle=0,end angle=280,radius=0.15];
        \draw[dashed, red] (1.25,-0.51) -- (2.35,-0.81); 
        \draw[<->, red]  (1.25,-0.41) arc[start angle=90,end angle=-70,radius=0.15];
        \draw[dashed, red] (2.00,-1.11) -- (3.15,-1.31); 
        \draw[<->, red]  (2.20,-1.11) arc[start angle=0,end angle=280,radius=0.15];
        \draw[dashed, red] (2.95,-1.11) -- (3.60,-0.81); 
        \draw[<->, red]  (3.15,-1.11) arc[start angle=0,end angle=240,radius=0.15];

        \draw[-latex, dashed,red] (-3.85,-2) -- node[right]{{\scriptsize{$\mathbf{p}_{BS}$}}}(-1.85,0);
    \end{tikzpicture}
    }
    \vspace*{-0.45em}
    \caption{Illustration of two~\ac{MRAV} configurations along with the global ($\mathcal{F}_W$) and untilted ($\mathcal{F}_U$) reference systems: under-actuated (left) and omnidirectional (right)~\cite{Aboudorra2023JINT}. Arcs represent the rotation direction of servos used for varying the thrust vector.}
    \label{fig:mrav}
\end{figure}

In contrast, omnidirectional \acp{MRAV} overcome these limitations by enabling simultaneous control over position and orientation \cite{Allenspach2020IJRR, Aboudorra2023JINT}. This is achieved by actively tilting their propellers using servo motors, optimizing energy consumption by aligning propeller spinning directions with the required orientation \cite{RyllTCST2015, Ryll2022TM}. This balance between dexterity and energy efficiency supports various task-specific demands and can effectively counteract challenges like jamming attacks or eavesdropping \cite{BonillaICASSP2024}. Figure \ref{fig:mrav} illustrates both under-actuated and omnidirectional configurations of an \ac{MRAV}, illustrating the versatility of these platforms.

The present study expands on the capabilities of omnidirectional \acp{MRAV} by exploring their use in scenarios involving eavesdroppers. This paper introduces an innovative approach where a team of omnidirectional \acp{MRAV}, one acting as an aerial \acl{BS} and the other as a friendly jammer, collaboratively work to secure communications for legitimate users while disrupting nearby eavesdroppers. This strategy, which leverages advanced motion planning techniques, represents a novel application to enhance the security of \ac{MRAV} communications in the presence of eavesdroppers.

The focus of this research is the joint optimization of both position and orientation (also known as pose) of omnidirectional \acp{MRAV}, marking the first instance, to the best of the authors' knowledge, of employing such a strategy to ensure secure communications in the presence of eavesdroppers. Previous research has demonstrated the use of \acp{MRAV} for secure communications, such as using a single \ac{MRAV} equipped with an antenna array to simultaneously support legitimate communications and jam eavesdroppers \cite{WuIEEETVT2020}. Another study \cite{ZhangIEEETWC2021} examined a dual-\ac{MRAV} network where one \ac{MRAV} collects data from users, and the other jams eavesdroppers, optimizing trajectories and power for maximum secrecy. In scenarios like \cite{LeeTVT2018}, \acp{MRAV} transmit private messages and jamming signals to protect communications, optimizing power and trajectory for the highest secrecy rates. Similarly, \cite{ZhongIEEECL2019} explores two \acp{MRAV} working together to transmit secure information to a ground node while jamming eavesdroppers. Other research, such as \cite{XuIEEETC2021}, considers the impact of environmental factors like wind on \ac{MRAV} stability and communication efficacy, optimizing systems against worst-case scenarios. However, none of the mentioned approaches address enhancing secrecy rates by physically orienting the antennas in such a way that jamming signals are directed at eavesdroppers and while information signals are directed at legitimate users, this is achieved by leveraging the omnidirectional capabilities and it is an alternative to the use of traditional beamforming. 

Our work builds on these studies by using a dual-\ac{MRAV} setup with cooperative jamming, utilizing omnidirectional \acp{MRAV}' ability to direct signals efficiently, thereby increasing secrecy rates and enhancing network security. This innovative application appears to be the first of its kind in utilizing omnidirectional \acp{MRAV} for improved network security. 



\section{System model}
\label{sec:systemModel}

We consider the scenario where a team of two~\acp{MRAV} is charged with providing secure and efficient communication to various ground users in the presence of $N$ eavesdroppers. One~\ac{MRAV} (\ac{MRAV}-I) communicates with the ground nodes while the other~\ac{MRAV} (\ac{MRAV}-J) protects the privacy of the communication by jamming eavesdroppers in the region.



\subsection{Eavesdroppers and legitimate users}
\label{sec:eavesdroppersLegitimateUsers}

We assume that~\ac{MRAV}-I transmits information to legitimate stationary nodes, and this transmission is vulnerable to interception by malicious eavesdroppers. To address this, the team of \acp{MRAV} handles one legitimate node at a time using
\ac{TDMA}\footnote{We assume the time slots are sufficiently long to allow the team of~\acp{MRAV} to position themselves optimally for each legitimate node and transmit data for an extended duration before moving to the next legitimate user.}. 
Therefore, we focus on the system during a single time slot, denoting $S_0$ as the active legitimate user at that time, and we will denote $\{S_j\}_{j=1}^N$ as the set of eavesdroppers. We represent the coordinates in the global reference frame $\mathcal{F}_W = \{O_W, \mathbf{x}_W, \mathbf{y}_W, \mathbf{z}_W\}$ of the node $S_j$ as $\mathbf{p}_{S_j} \in \mathbb{R}^3$ (see Figure~\ref{fig:mrav}). Without loss of generality, we assume that $\mathbf{p}_{S_0}=\mathbf{0}$, with $\mathbf{0} \in \mathbb{R}^3$.  We also assume that all eavesdroppers and legitimate users are equipped with single omnidirectional antennas. 
To ensure secure transmission between \ac{MRAV}-I and the user $S_0$, during its allotted time slot, the \ac{MRAV}-J emits artificial noise to disrupt the eavesdroppers.



\subsection{Omnidirectional \acp{MRAV}}
\label{sec:omnidirectionalMRAV}

We consider both~\acp{MRAV} are omnidirectional. The position of the \ac{MRAV}-$i$ ($i\in\{I,J\}$) in the global reference frame $\mathcal{F}_W$ is denoted as $\mathbf{p}_{U_i} \in \mathbb{R}^3$. Additionally, we introduce an $i$-th untilted coordinate frame $\mathcal{F}_{U_i} = \{O_{U_i}, \mathbf{x}_{U_i}, \mathbf{y}_{U_i}, \mathbf{z}_{U_i}\}$, which is aligned with the global coordinate frame $\mathcal{F}_W$ and centered at $\mathbf{p}_{U_i}$ (see Figure~\ref{fig:mrav}). To precisely describe the orientation of \ac{MRAV}-$i$ in the global coordinate frame $\mathcal{F}_W$, we use Euler angles, specifically roll ($\varphi_i$), pitch ($\vartheta_i$), and yaw ($\psi_i$). We refer to the orientation of \ac{MRAV}-$i$ as $\bm{\eta}_{i}=[\varphi_i, \vartheta_i, \psi_i]^\top \in \mathbb{R}^3$.

The omnidirectional~\ac{MRAV}-$i$ is equipped with a single antenna, which is located at $\mathbf{p}_{U_i}$, on its upper surface and oriented according to the following vector expressed in $\mathcal{F}_{U_i}$:
\begin{equation}\label{eq:1.1}
\bm{\Upsilon}_i(\bm{\eta}_{i})= 
    \begin{bmatrix}
        \cos(\varphi_i)\sin(\vartheta_i)\cos(\psi_i)+\sin(\varphi_i)\sin(\psi_i)\\
        \cos(\varphi_i)\sin(\vartheta_i)\sin(\psi_i)-\sin(\varphi_i)\cos(\psi_i)\\
        \cos(\varphi_i)\cos(\vartheta_i)
   \end{bmatrix}.
\end{equation}

We also consider that the~\acp{MRAV} experience slight jittering due to the wind and other control issues. To account for this jittering, we follow the same model as \cite{ChenIEEETWC2024} which adds the jittering to the orientation vector:
%
    $\bm{\eta}_{i}=\bm{\eta}_{i}^d+\bm{w}_i(t),$
%
where $\bm{\eta}_{i}^d\in\mathbb{R}^3$ represents the desired Euler angles for the \ac{MRAV}-$i$, and $\bm{w}_i(t)\in\mathbb{R}^3$ is a zero-mean Gaussian random process with covariance matrix $\mathbf{C}_i=\sigma_{U_i}^2\mathbf{I}_{3}$, where $\mathbf{I}_3 \in \mathbb{R}^{3 \times 3}$ is the identity matrix and $\sigma_{U_i}^2$ is the variance. For notational simplicity we assume the same variance for all three Euler angles, note that this simplification does not affect our proposed method. We write the joint \ac{PDF} of the three components of $\bm{w}_i(t)$ as %
    $f_{w_i}({\varphi_i}, \vartheta_i, \psi_{i}) = g({\varphi_i}; \sigma_{U_i}^2) g({\vartheta_i}; \sigma_{U_i}^2) g({\psi_i}; \sigma_{U_i}^2)$,
%
where $g(\cdot;\sigma_{U_i}^2)$ is a Gaussian distribution of zero-mean and variance $\sigma^2_{U_i}$. We also assume that $\bm{w}_I(t)$ and $\bm{w}_J(t)$ are statistically independent and identically distributed random variables.  



\subsection{Channel model}
\label{sec:channelModel}

We consider both~\acp{MRAV}' antennas to be dipoles. As a result, the normalized power radiation patterns of the information and jammer~\acp{MRAV} can be characterized as follows~\cite{MIRON20069}:
\begin{equation}\label{eq:1.2}   
G(\gamma) = \sin^2(\gamma), 
\end{equation}
where $\gamma$ is the elevation angle component of the \ac{AoD} of the antenna in consideration. Now, let us consider the communications link between \ac{MRAV}-$i$ and node $S_j$. The cosine of the elevation angle component for this communication link can be expressed as:
\begin{eqnarray}\label{eq:1.3}
  \cos(\gamma_{ij}) = \left\langle \frac{\mathbf{p}_{S_j} -\mathbf{p}_{U_i}}{\|\mathbf{p}_{S_j} - \mathbf{p}_{U_i}\|}, \bm{\Upsilon}(\bm{\eta}_i)\right\rangle, 
\end{eqnarray}
where $\langle\cdot,\cdot\rangle$ represents the inner product operation, and $\gamma_{ij}$ is the elevation angle of the~\ac{AoD} between \ac{MRAV}-$i$ and node $S_j$. Subsequently, we can formulate the radiation pattern of the antenna of \ac{MRAV}-$i$ as:
\begin{eqnarray}\label{eq:1.4}
    G(\gamma_{ij}) = 1-\left\langle \frac{\mathbf{p}_{S_j}-\mathbf{p}_{U_i}}{\|\mathbf{p}_{S_j}-\mathbf{p}_{U_i}\|}, \bm{\Upsilon}(\bm{\eta}_{i})\right\rangle^2. 
\end{eqnarray}

We assume that the air-to-ground channel is mainly influenced by path loss \cite{XuIEEETC2021}. Additionally, we suppose that the \acp{MRAV} know the location of the legitimate node and have estimates of the locations of the eavesdroppers\footnote{The locations of the eavesdroppers can be estimated, for instance, as described in \cite{MukherjeeICASSP2012} and then shared with the \acp{MRAV}.};  
we denote those estimates as $\{\hat{\mathbf{p}}_{S_j}\}_{j=1}^N$. We also assume that the \acp{MRAV} do not know the variance of the estimates nor their statistical distributions. The distance between~\ac{MRAV}-$i$ and node $S_j$ is  $d_{ij}\triangleq\|\mathbf{p}_{S_j} - \mathbf{p}_{U_i}\|$, and the~\ac{SINR} at node $S_j$ is:
\begin{equation}\label{eq:1.5}
    \Gamma_j = \frac{G(\gamma_{Ij}) P_I d_{Ij}^{-2}}{G(\gamma_{Jj}) P_J d_{Jj}^{-2} + \sigma_j^2}, \; \forall  j \in \{0,\dots,N\},
\end{equation}
where $P_i$ is the transmit power of \ac{MRAV}-$i$, and $\sigma_j^2$ is the variance of additive white Gaussian noise of node $j$. Finally, the secrecy rate, calculated with the estimated positions of the eavesdroppers, is: 
\begin{equation}\label{eq:1.6}
    \resizebox{0.91\hsize}{!}{$
    R(\bm{\Gamma})=B\left[ \log_2(1+\Gamma_0)- \max\limits_{j\in \{1,\dots,N\} }(\log_2(1+\hat{\Gamma}_j))\right]^+,
    $}
\end{equation}
where $[\cdot]^+=\max(0,\cdot)$, $B$ represents the allocated bandwidth, $\bm{\Gamma}=[\Gamma_0,\, \hat{\Gamma}_1,\,\dots,\,\hat{\Gamma}_N ]^\top$, and $\{\hat{\Gamma}_{j}\}_{j=1}^N$ are the estimated \ac{SINR} at the eavesdroppers obtained by using the estimated positions $\{\hat{\mathbf{p}}_{S_j}\}_{j=1}^N$ instead of the real ones $\{{\mathbf{p}}_{S_j}\}_{j=1}^N$. 



\section{Problem Formulation}
\label{sec:problemFormulation}

We aim to optimize the desired pose (position and orientation) of the information \ac{MRAV} and the jamming \ac{MRAV}, along with their transmission powers, to maximize the secrecy rate of the legitimate user $S_0$. This problem is stated as follows:
\begin{subequations} \label{eq:2.2a}
    \begin{align}
    &\maximize_{\bm{\eta}_I^d, \, \mathbf{p}_{U_I}, \bm{\eta}_J^d, \, \mathbf{p}_{U_J}, P_I, P_J} \mathbb{E}[R(\bm{\Gamma})] \label{subeq:2.2a_objective}\\
    &\qquad \;\, \text{s.t.}~\quad \underline{z} \leq \mathbf{e}_3^\top \mathbf{p}_{U_i} \leq \bar{z}, \label{subeq:2.2a_first}\\
    &\qquad \qquad \;\;\, \left\|
     \begin{bmatrix}
        \mathbf{I}_2 & \\
        & 0 \\
    \end{bmatrix} \bm{\eta}_i \right\|_{\infty} \leq \frac{\pi}{2}, \label{subeq:2.2a_second} \\
    &\qquad \qquad \;\;\;\, 0 \leq P_i \leq \bar{P}, \label{subeq:2.2a_fourth} \\
    &\qquad \qquad \;\;\;\, \text{with} \; \psi_i=0, \; \forall i \in \{I, J\}. \label{subeq:2.2a_third}
    \end{align}
\end{subequations}
In this formulation, the objective function \eqref{subeq:2.2a_objective} represents the expected value (accounting for disturbances in the~\acp{MRAV} orientation) of the secrecy rate for the user $S_0$ (see \eqref{eq:1.6}):
\begin{equation}\label{eq:2.2b}
    \resizebox{0.91\hsize}{!}{$
    \mathbb{E}[R(\bm{\Gamma})] = \mathlarger{\iiiint_{\mathbb{R}^4}} R(\bm{\Gamma}) \prod\limits_{i=\{J,I\}} g({\varphi_i}; \sigma_{U_i}^2) g({\vartheta_i}; \sigma_{U_i}^2) \mathrm{d}{\varphi_i} \mathrm{d}{\vartheta_i}.
    $}
\end{equation}
Constraint \eqref{subeq:2.2a_first} restricts the altitude of the \acp{MRAV} within specified limits, denoted as $\underline{z}$ and $\bar{z}$. Here, $\mathbf{e}_3=(0,0,1)^\top$ corresponds the third column of the identity matrix $\mathbf{I}_3$. Constraint \eqref{subeq:2.2a_second} defines permissible ranges for the pitch ($\vartheta$) and roll ($\varphi$) angles, with $\mathbf{I}_2 \in \mathbb{R}^{2 \times 2}$. The optimization problem explores a search space with twelve dimensions, encompassing the position of both \acp{MRAV} $\{\mathbf{p}_{U_i}\}_{i=\{J,I\}}$, their roll and pitch angles $\{\varphi_i,\vartheta_i\}_{i=\{J,I\}}$ as well as their transmission powers $\{P_i\}_{i=\{J,I\}}$. Notably, we omit consideration of the yaw angle ($\psi$) \eqref{subeq:2.2a_third} due to the omnidirectional radiation pattern of the antenna, which exhibits uniformity around its axis. Lastly, constraint \eqref{subeq:2.2a_fourth} enforces that the transmission powers $P_i$ remain strictly positive and do not exceed a maximum value denoted as $\bar{P}$.

The resulting optimization problem \eqref{eq:2.2a} is non-convex and features non-smooth and nonlinear functions, such as the $[\cdot]^+$ and $\max(\cdot)$ functions in \eqref{eq:1.6}. This makes it challenging to find a global optimal solution. Therefore, we will introduce a novel approach to decompose the problem \eqref{eq:2.2a} into simpler subproblems, ultimately deriving a suboptimal solution.



\section{Solution}
\label{sec:solution}

In the proposed approach, we tackle the optimization problem in four distinct phases which are executed iteratively, and they are described individually in the next subsections.

\subsection{Phase 1: \ac{MRAV}-I orientation}
\label{sec:phase1}

In this section, we introduce the concept of the \textit{maximum gain plane} for the \ac{MRAV}-$i$, with $i \in \{I, J\}$, and explain how it is used to determine the orientation of the \ac{MRAV}-I. This plane, denoted as $\mathcal{X}_i$, is defined as follows:
%
 $   \mathcal{X}_i=\{\mathbf{q}+\mathbf{p}_{U_i} : \, \mathbf{q}\times\bm{\Upsilon}(\bm{\eta}_i^d)=0, \mathbf{q}\in\mathbb{R}^3\}.$
This plane represents the region where the antenna of \ac{MRAV}-$i$ achieves maximum gain, assuming the orientation is set to the desired one, $\bm{\eta}_i^d$. Additionally, we define the \textit{ground plane} as $\mathcal{G}$, consisting of points in $\mathcal{F}_W$ with zero height above the ground:
%
    $\mathcal{G}=\{[x,y,0]^\top: x,y\in\mathbb{R}\}.$
Before determining the \ac{MRAV}-I orientation, we introduce a line denoted as $\mathcal{L}_i$. This line is the intersection of $\mathcal{X}_i$ and $\mathcal{G}$ (i.e., $    \mathcal{L}_i=\mathcal{X}_i\cap\mathcal{G}.$) and is determined based on the position of \ac{MRAV}-I ($\mathbf{p}_{U_I}$).

\begin{figure}[tb]
    \centering
    \begin{tikzpicture}
        \draw[-latex] (-7.45,0.2) -- (-7.45,0.7) node[above]{{\scriptsize{$\mathbf{z}_W$}}}; 
        \draw[-latex] (-7.45,0.2) -- (-6.85,-0.09) node[above]{{\scriptsize{$\mathbf{y}_W$}}}; 
        \draw[-latex] (-7.45,0.2) -- (-8.15,0.09) node[above]{{\scriptsize{$\mathbf{x}_W$}}}; 
        \node at (-7.45,0.2) [below]{{\scriptsize{$O_W$}}};

        \draw[line width=1.25pt] (-4.45,0.2) -- (-3.85,0.55);
        \draw (-3.6,0.5) node[above]{{\scriptsize{$\bm{\Upsilon}_I(\bm{\eta}_I^d)$}}};
        \draw (-2.2,0) node[above]{{\scriptsize{antenna \ac{MRAV}-I}}};
        \draw[-latex, densely dotted] (-3.1,0.2) -- (-3.9,0.45);

        \draw (-3.95,0.55) node[left]{{\scriptsize{$\mathbf{z}_{U_I}$}}}; 
        \draw[-latex] (-4.45,0.2) -- (-3.85,-0.43);
        \draw (-3.70,-0.43) node[above]{{\scriptsize{$\mathbf{y}_{U_I}$}}}; 
        \draw[-latex] (-4.45,0.2) -- (-5.15,0.2) node[above]{{\scriptsize{$\mathbf{x}_{U_I}$}}}; 
        \node at (-4.45,0.2) [below]{{\scriptsize{$O_{U_I}$}}};
        
        \draw[densely dash dot] (-4.25, 0.32) -- (-4.05, 0.12) -- (-4.25,-0.02); 
        \draw[densely dash dot] (-4.25, 0.32) -- (-4.40, 0.05) -- (-4.50,-0.02);

        \draw[-latex] (-3.85,-2) -- (-3.85,-1.5) node[above]{{\scriptsize{$z$}}}; 
        \draw[-latex] (-3.85,-2) -- (-3.25,-2.29) node[above]{{\scriptsize{$y$}}}; 
        \draw[-latex] (-3.85,-2) -- (-4.55,-2.19) node[below]{{\scriptsize{$x$}}}; 
        \node at (-3.85,-2) [below]{{\scriptsize{$O$}}};

        \draw[dashed] (-6.25,-2.4)  -- (-3.85,-2) -- (-1.45,-1.6); 
        \draw (-6.25,-2.4) node[above]{{\scriptsize{$\mathcal{L}_I$}}};

        \draw[-latex, dash dot] (-4.45,0.2) -- (-5.5,-2.26) node[below]{{\scriptsize{$l_1$}}};
        \draw (-5.5,-2.28) node[fill=black, circle, minimum size = 0.075cm, inner sep=0pt]{};
        \draw[-latex, dash dot] (-4.45,0.2) -- (-2.6,-1.76) node[below]{{\scriptsize{$l_2$}}};
        \draw (-2.6,-1.78) node[fill=black, circle, minimum size = 0.075cm, inner sep=0pt]{};

        \draw (-5.5,-3.28) node[fill=white, draw, circle, minimum size = 0.075cm, inner sep=0pt]{};
        \draw (-5.5,-3.28) node[above]{{\scriptsize{eavesdropper\textsubscript{2}}}};
        \draw (-2.5,-2.28) node[fill=white, draw, circle, minimum size = 0.075cm, inner sep=0pt]{};
        \draw (-2.5,-2.28) node[below]{{\scriptsize{eavesdropper\textsubscript{1}}}};
        \draw (-4.125,-1.20) node[fill=white, draw, circle, minimum size = 0.075cm, inner sep=0pt]{};
        \draw (-4.125,-1.20) node[above]{{\scriptsize{eavesdropper\textsubscript{3}}}};
        
    \end{tikzpicture}
    \vspace{-0.35em}
    \caption{A schematic representation of the $\mathcal{X}_i$ and $\mathcal{G}_i$ planes along with the eavesdroppers, the legitimate user (located at the origin $O$), and the \ac{MRAV}-I.}
    \label{fig:overallScenarioPhase1}
\end{figure}
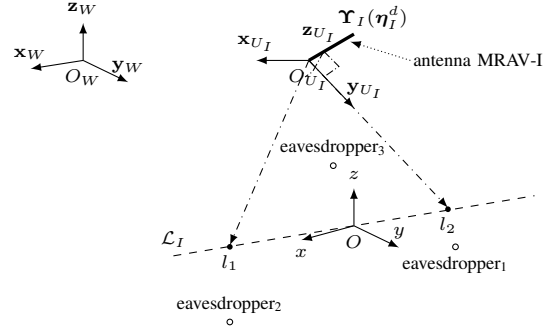

Figure \ref{fig:overallScenarioPhase1} depicts the overall scenario, and the orientation of the \ac{MRAV}-I is chosen to satisfy two criteria:
\begin{itemize}
    \item \displayCriterion[CR1]{Maximum antenna gain} Ensure that the maximum antenna gain of the \ac{MRAV}-I is directed towards the legitimate user, i.e., $\mathbf{p}_{S_0} \in \mathcal{L}_I$.
    
    \item \displayCriterion[CR2]{Horizontal distance} Maximize the horizontal distance between $\mathcal{L}_I$ and the eavesdroppers to minimize the gain of \ac{MRAV}-I's antenna observed by the eavesdroppers.
\end{itemize}  

The line $\mathcal{L}_I$ can be represented as the set:
%
    $\mathcal{L}_I=\{[x,\, y,\, 0]^\top \,: y=ax, x\in\mathbb{R}\},$
%
with $a \in \mathbb{R}$ being a parameter to be optimized. To find the optimal $a$, we define a vector $\mathbf{w}$ that is orthonormal to $\mathcal{L}_I$:
%
    $\mathbf{w}=[ -a, 1,\ 0\ ]^\top \left(\sqrt{1+a^2}\right)^{-1}.$

Given that $\mathbf{p}_{S_0}=\mathbf{0}$, we already satisfy \refCriterion[CR1] regardless the value of $a$. To satisfy \refCriterion[CR2], we optimize the parameter $a$ to maximize the minimum distance between the eavesdroppers and the line $\mathcal{L}_I$. The minimum distance, denoted as $H_i$, between $S_i$ and $\mathcal{L}_I$, is given by:
%
   $H_i=|\mathbf{w}^\top \mathbf{p}_{S_i}|.$

Thus, we formulate the optimization problem as maximizing the minimum value of $\hat{H}_i^2=|\mathbf{w}^\top \hat{\mathbf{p}}_{S_i}|^2$ by varying the parameter $a$ of the vector $\mathbf{w}$. To avoid numerical issues associated with an unbounded search space for $a$, we use the parameter $\nu$ defined as $a=\tan(\nu)$ with $\nu\in[-\pi/2,+\pi/2]$. Note that this implies that $\nu=\pi/2$ and $\nu=-\pi/2$ both correspond to a vertical line and are treated equivalently. Therefore, the optimization problem becomes\footnote{Note that we are using the estimated locations of the eavesdroppers.}:
\begin{equation}\label{eq:opA}
    \maximize_{a} \Biggl( \min_{j=\{1,\dots,N\}} \hat{H}_j^2 \Biggr).
\end{equation}

The optimization problem \eqref{eq:opA} is non-convex and can have multiple local maxima and minima. The number of local maxima and minima are proportional to the number of eavesdroppers $N$ and also depends on their angular locations relative to $S_0$. To address this, we can employ numerical optimization techniques such as simulated annealing, which can handle non-convex problems with multiple local optima efficiently when the search space is bounded. Once we have determined $\mathcal{L}_I$, we can calculate the desired orientation of \ac{MRAV}-I, as:
\begin{equation}\label{eq:4.7}
    \resizebox{0.89\hsize}{!}{$
    \bm{\eta}_I^d = \left[
    \begin{array}{c}
       -\mathrm{asin}\left([\bm{\Upsilon}(\bm{\eta}_I^d)]_y\right)    \\
       \arctan2\left(\frac{[\bm{\Upsilon}(\bm{\eta}_I^d)]_x}{\cos\left(\mathrm{asin}\left([\bm{\Upsilon}(\bm{\eta}_I^d)]_y\right)\right)},\frac{[\bm{\Upsilon}(\bm{\eta}_I^d)]_z}{\cos\left(\mathrm{asin}\left([\bm{\Upsilon}(\bm{\eta}_I^d)]_y\right)\right)}\right)    \\
      0    
    \end{array}
    \right],
    $}
\end{equation}
%
%
where $\mathrm{asin}(\cdot)$ is the arcsine function, $\arctan2(x,y)$ is the four quadrants inverse tangent, and: 
\begin{equation}\label{eq:4.8}
    \bm{\Upsilon}(\bm{\eta}_I^d) = \left( \frac{ \ell_1 -\mathbf{p}_{U_I} }{ |\ell_1 - \mathbf{p}_{U_I}|_2} \right ) \times \left( \frac{\ell_2 - \mathbf{p}_{U_I}}{ |\ell_2 - \mathbf{p}_{U_I}|_2} \right),
\end{equation}
where $\ell_1$ and $\ell_2$ are any two points in $\mathcal{L}_I$, and $[\bm{\Upsilon}(\bm{\eta}_I^d)]_x$,  $[\bm{\Upsilon}(\bm{\eta}_I^d)]_y$, and $[\bm{\Upsilon}(\bm{\eta}_I^d)]_z$ represent the $x$, $y$, and $z$ components of the vector $\bm{\Upsilon}(\bm{\eta}_I^d)$, respectively. This orientation ensures that the \ac{MRAV}-I provides maximum antenna gain in the direction of the legitimate user while providing low antenna gain in the direction of eavesdroppers.

\subsection{Phase 2: \ac{MRAV}-J orientation}
\label{sec:phase2}

Regarding the \ac{MRAV}-J orientation, we propose orienting it to minimize its interference to the legitimate user $S_0$: 
\begin{equation}\label{eq:4.9}
    \bm{\Upsilon}(\bm{\eta}_J^d)=\frac{\mathbf{p}_{S_0}-\mathbf{p}_{U_I}}{\|\mathbf{p}_{S_0}-\mathbf{p}_{U_I}\|_2}.
\end{equation} 
Then $\bm{\eta}_J^d$ is extracted from $\bm{\Upsilon}(\bm{\eta}_J^d)$ using the same formula (\ref{eq:4.7}) after replacing $\bm{\Upsilon}(\bm{\eta}_I^d)$ with $\bm{\Upsilon}(\bm{\eta}_J^d)$. If the \ac{MRAV}-J experienced no jittering, then $\bm{\Upsilon}(\bm{\eta}_J) = \bm{\Upsilon}(\bm{\eta}_J^d)$, and thus the null of the antenna of the \ac{MRAV}-J would be permanently oriented to the legitimate user. Consequently, the \ac{MRAV}-J could radiate as much power as possible without interfering with the receiver of the legitimate user. However, due to the jittering, the legitimate user will receive some small amount of power from the \ac{MRAV}-J, which will slightly degrade its \ac{SINR}.



\subsection{Phase 3: Transmission Power}
\label{sec:phase3}

With the orientations of both \ac{MRAV}-I and \ac{MRAV}-J determined as functions of their positions, we proceed by substituting \eqref{eq:4.8}--\eqref{eq:4.9} expressions into the original optimization problem \eqref{eq:2.2a}. The resultant problem is formulated as follows:
\begin{subequations} \label{eq:phase3}
    \begin{align}
    &\maximize_{\mathbf{p}_{U_I}, \mathbf{p}_{U_J}, P_I, P_J} \mathbb{E}[R(\bm{\Gamma})] \\
    &\quad \;\, \text{s.t.}~\quad \underline{z} \leq \mathbf{e}_3^\top \mathbf{p}_{U_i} \leq \bar{z}, \\
    &\quad \qquad \;\;\;\, 0 \leq P_i \leq \bar{P},\\
    &\quad \qquad \;\;\;\, (\ref{eq:4.8}),(\ref{eq:4.9}).
    \end{align}
\end{subequations}
It should be noted that directly optimizing the secrecy rate presents challenges due to the flat valleys introduced by the nonlinear function $[\cdot]^+ = \max(0, \cdot)$ in $R(\bm{\Gamma})$. To mitigate this issue, the utility function is adjusted by replacing $R(\bm{\Gamma})$ with:
\begin{equation} \label{eq:phase3b}
    \resizebox{0.89\hsize}{!}{$
    \hat{R}(\bm{\Gamma}) = B\left[ \log_2(1+\Gamma_0) - \max\limits_{j\in \{1, \dots, N\}} (\log_2(1+\hat{\Gamma}_j)) \right].
    $}
\end{equation}
Additionally, computing the fourth-dimensional integral \eqref{eq:2.2a} numerically is computationally demanding. Therefore, instead of numerically evaluating the integral to obtain the expected value over the jittering, a Monte Carlo approach is adopted. Various realizations of the Euler angles ($\bm{\eta}$) for both \acp{MRAV} are generated according to the Gaussian model considered for the jittering, the secrecy rate is computed for each realization, and then averaged. By generating a sufficiently large number of realizations, the mean closely approximates the expected value, making this method computationally efficient.

\subsection{Phase 4: Position Optimization}
\label{sec:phase4}

Upon computing the expression for the secrecy rate as outlined in Section \ref{sec:phase3}, a numerical method is employed to solve the optimization problem concerning the 3D positions of both \acp{MRAV} and their transmission powers. Specifically, an interior point search method \cite{boyd2004convex} is utilized to efficiently handle the constrained optimization \eqref{eq:phase3} with the modified optimization target \eqref{eq:phase3b}. Once the optimal positions and powers are obtained, they are reintroduced into the initial problem \eqref{eq:2.2a}, as described in Sections \ref{sec:phase1} and \ref{sec:phase2}, to determine the optimal orientations for the updated positions and powers. This iterative process continues until a predefined number of iterations is reached or convergence is achieved.

\section{Simulations}
\label{sec:simulations}

\begin{figure*}[h]
    \hspace*{2.5em}
    \begin{subfigure}{0.6\columnwidth}
        \centering
        \hspace*{-3.4em}
         \adjincludegraphics[width=1.2\columnwidth, trim={{0.05\width} {0.05\height} {0.0\width} {.02\height}}, clip, keepaspectratio]{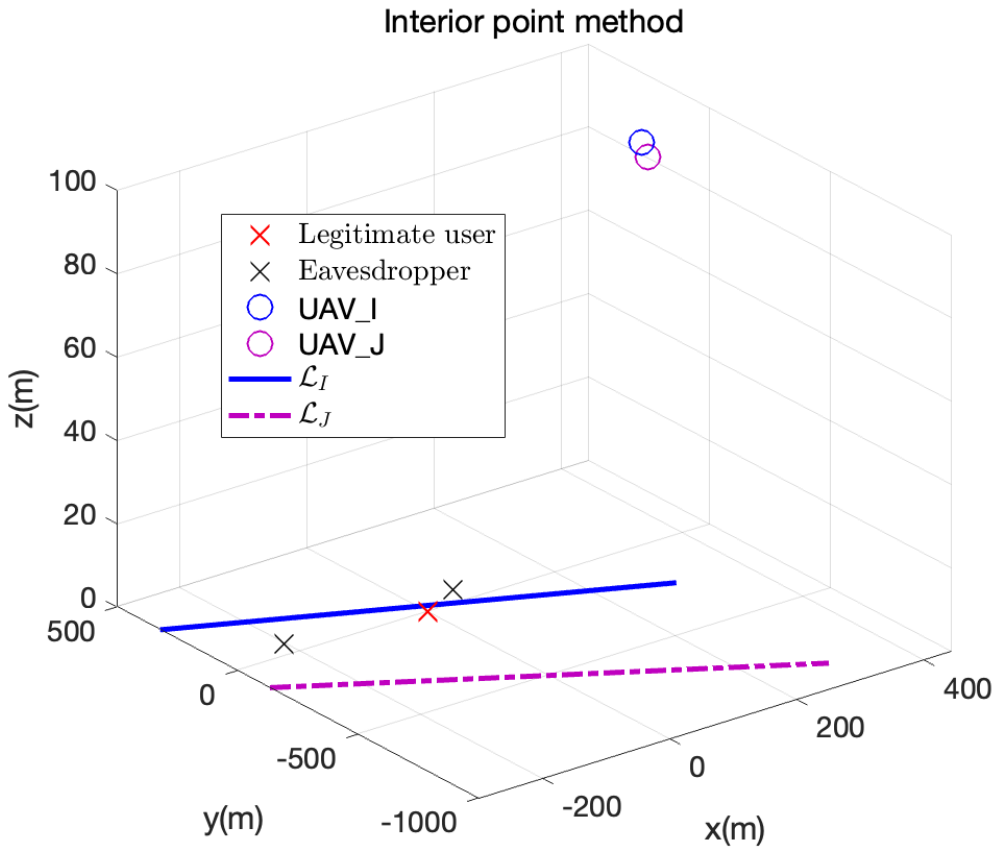}
         \vspace*{-5.5em}
         \caption{Interior Point.}
         \label{fig:interiorPoint}
    \end{subfigure}
    \hspace*{-1.9em}
    \begin{subfigure}{0.6\columnwidth}
        \centering
        \hspace*{0.5em}
        \adjincludegraphics[width=1.2\columnwidth, trim={{0.05\width} {0.05\height} {0.0\width} {.02\height}}, clip, keepaspectratio]{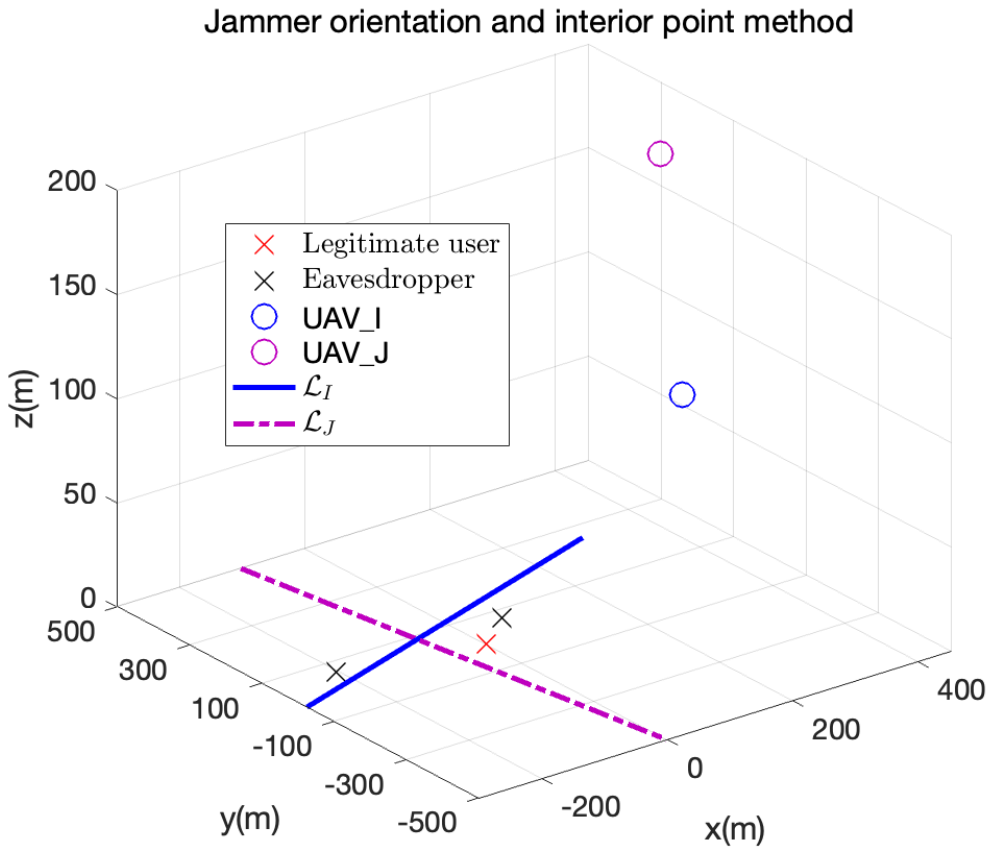} 
        \vspace*{-5.5em}
        \caption{Jammer Orientation.}
        \label{fig:jammerOrientation}
    \end{subfigure}
    \hspace*{0.50em}
    \begin{subfigure}{0.6\columnwidth}
        \hspace*{1.0em}
        \centering
        \adjincludegraphics[width=1.2\columnwidth, trim={{0.055\width} {0.05\height} {0.0\width} {.02\height}}, clip, keepaspectratio]{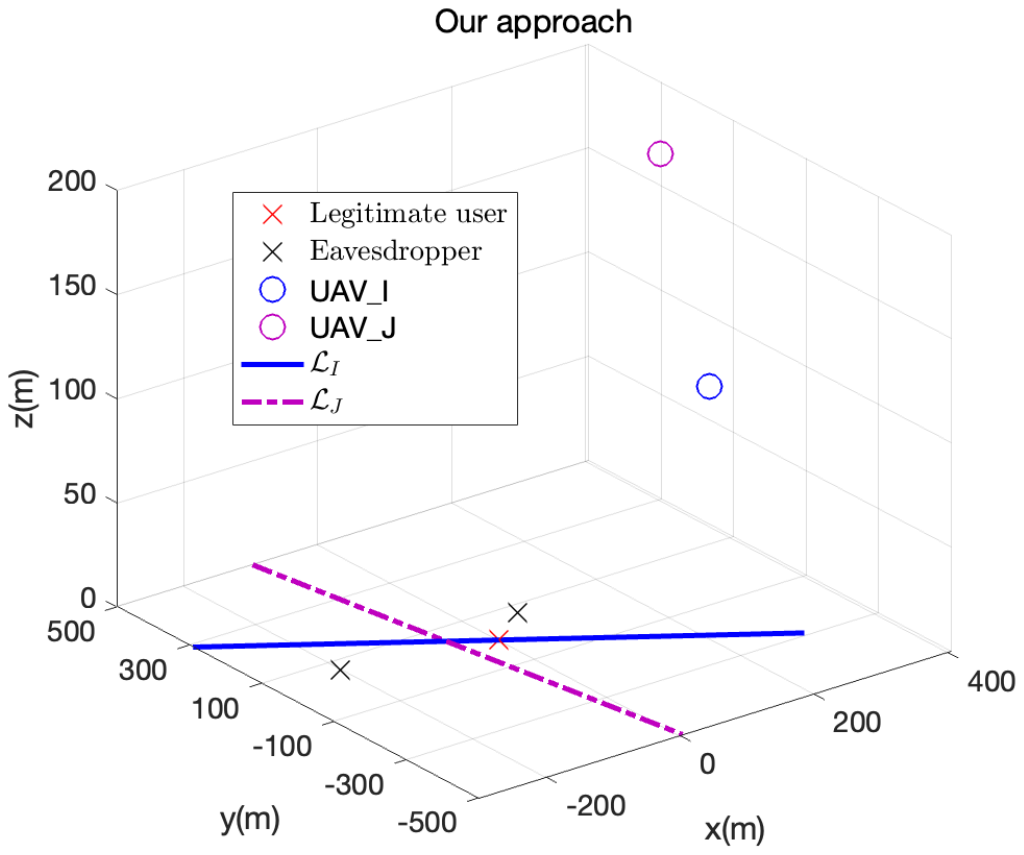} 
        \vspace*{-5.5em}
        \caption{Our Approach.}
        \label{fig:ourApproach}
    \end{subfigure}
    \vspace*{-0.45em}
    \caption{Final 3D positions obtained using the interior-point method, combined jammer orientation and interior-point method, and our proposed approach for scenarios with 2 eavesdroppers.}
    \label{fig3}
\end{figure*}

%


In this section, we evaluate the performance of our technique under different conditions. Firstly, we examine the scenario where the \acp{MRAV} operate without experiencing any jitter, followed by an assessment of the performance of our suboptimal solution (see Sections \ref{sec:phase3} and \ref{sec:phase4}). Additionally, we compare the effectiveness of the omnidirectional \acp{MRAV} with under-actuated ones in this context. Finally, we investigate the impact of jittering on the system's performance.




\subsection{Simulation Setup}
\label{sec:simulationSetup}

To evaluate our proposed approach, we consider an area of $\SI{1000}{\meter} \times \SI{1000}{\meter}$ with a legitimate user positioned at the origin. We study the scenario where $2$ eavesdroppers are randomly distributed within this area. Both the information and jammer \acp{MRAV} are initially placed within the region, with altitudes ranging from $\SIrange{80}{300}{\meter}$ ($\underline{z}$ and $\bar{z}$).
We assume that the \acp{MRAV} transmit with a maximum power $\bar{P}$ within the range of $\bar{P} \in [0.1, 1.1]\,\si{\watt}$. The noise variance for both legitimate users and eavesdroppers is set to $\sigma_l = \sigma_e = 10^{-12}\,\si{\watt}$. The allocated bandwidth for each \ac{MRAV} is $B = \SI{1}{\mega\hertz}$.

To assess our approach, we compare it against the following  benchmarks:
\textit{1) Interior-point method}: Solves the optimization problem described in \eqref{eq:2.2a} \ac{wrt} all optimization variables, including the orientations, positions, and powers of the \acp{MRAV}, using the interior-point method. \textit{2) Conventional \ac{MRAV}}: Assumes under-actuated \acp{MRAV} with fixed orientation and employs the interior-point method to optimize the \acp{MRAV} positions and transmit powers. 




\subsection{Simulation Results}
\label{sec:simulationResults}

\begin{figure*}[h]
    \centering
    \begin{subfigure}{0.6\columnwidth}
        \centering
         \adjincludegraphics[width=1.125\columnwidth, trim={{.1\width} {.2\height} {.05\width} {.2\height}}, clip, keepaspectratio]{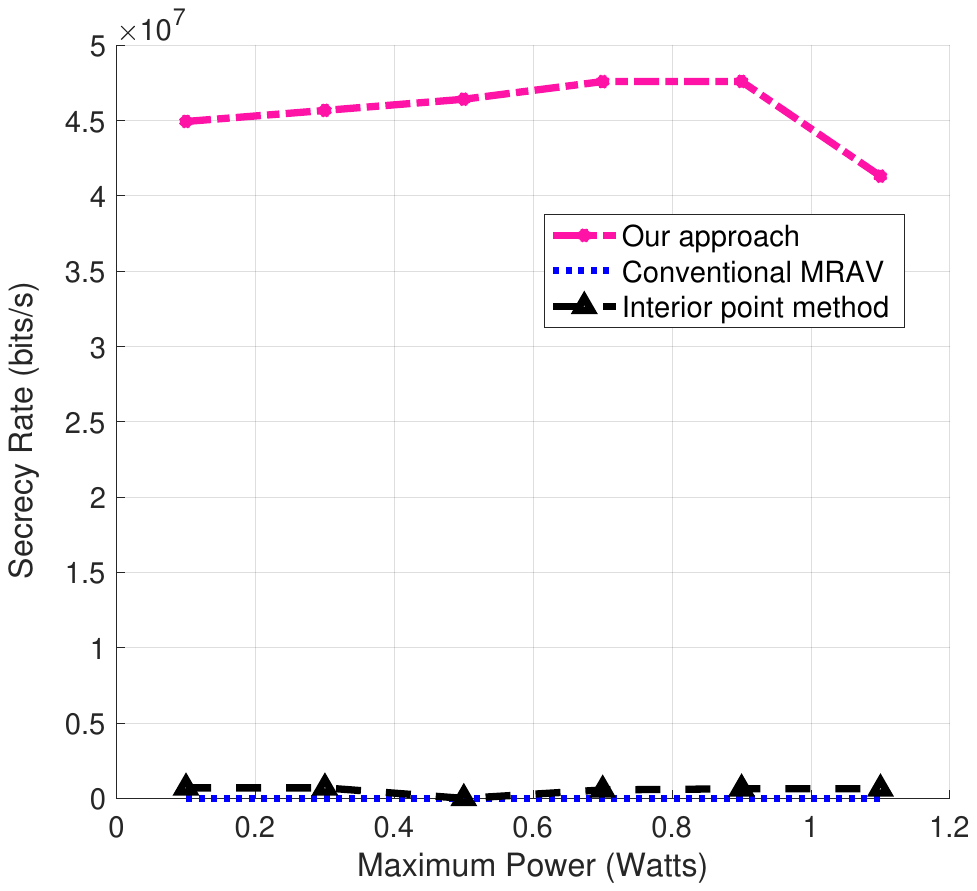}
         \vspace*{-1.85em}
         \caption{Without jittering and uncertainty.{\textcolor{white}{about the eavesdropper's position}}}
         \label{fig:withouJittering}
    \end{subfigure}
    \hspace*{2em}
    \begin{subfigure}{0.6\columnwidth}
        \centering
         \adjincludegraphics[width=1.125\columnwidth, trim={{.1\width} {.2\height} {.05\width} {.2\height}}, clip, keepaspectratio]{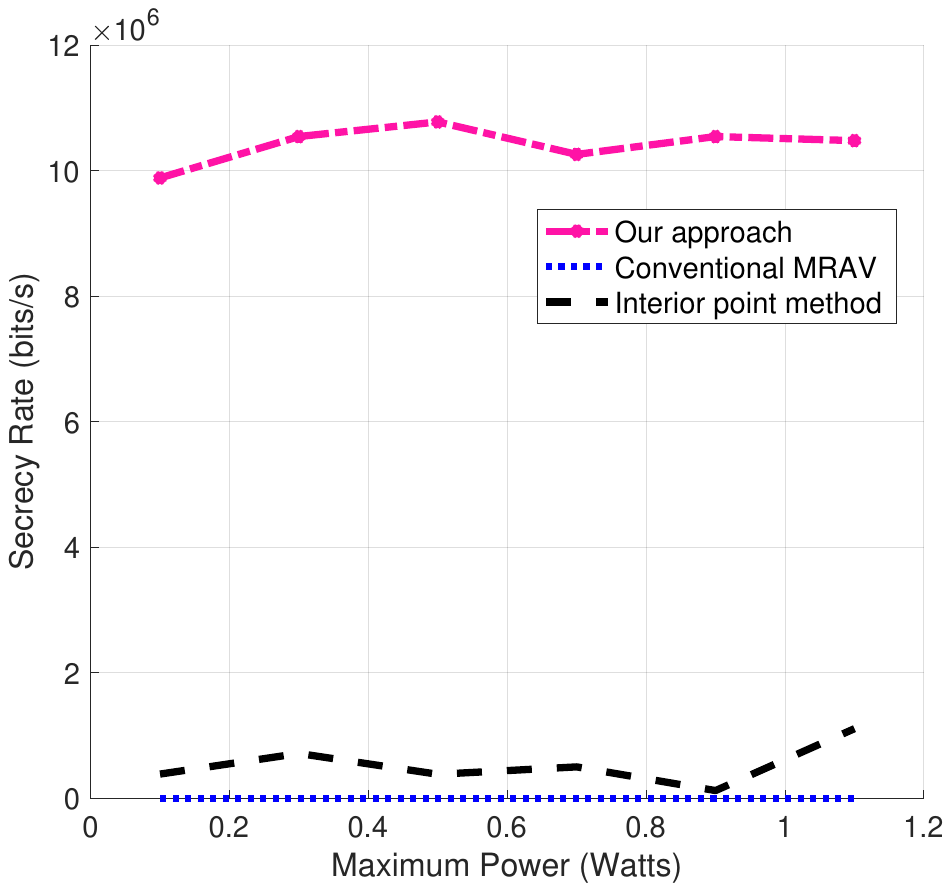}
         \vspace*{-1.85em}
         \caption{With jittering.{\textcolor{white}{ about the eavesdropper's position}}}
         \label{fig:withJittering}
    \end{subfigure}
    \hspace*{2em}
    \begin{subfigure}{0.6\columnwidth}
        \centering
         \adjincludegraphics[width=1.18\columnwidth, trim={{.1\width} {.2\height} {.05\width} {.2\height}}, clip, keepaspectratio]{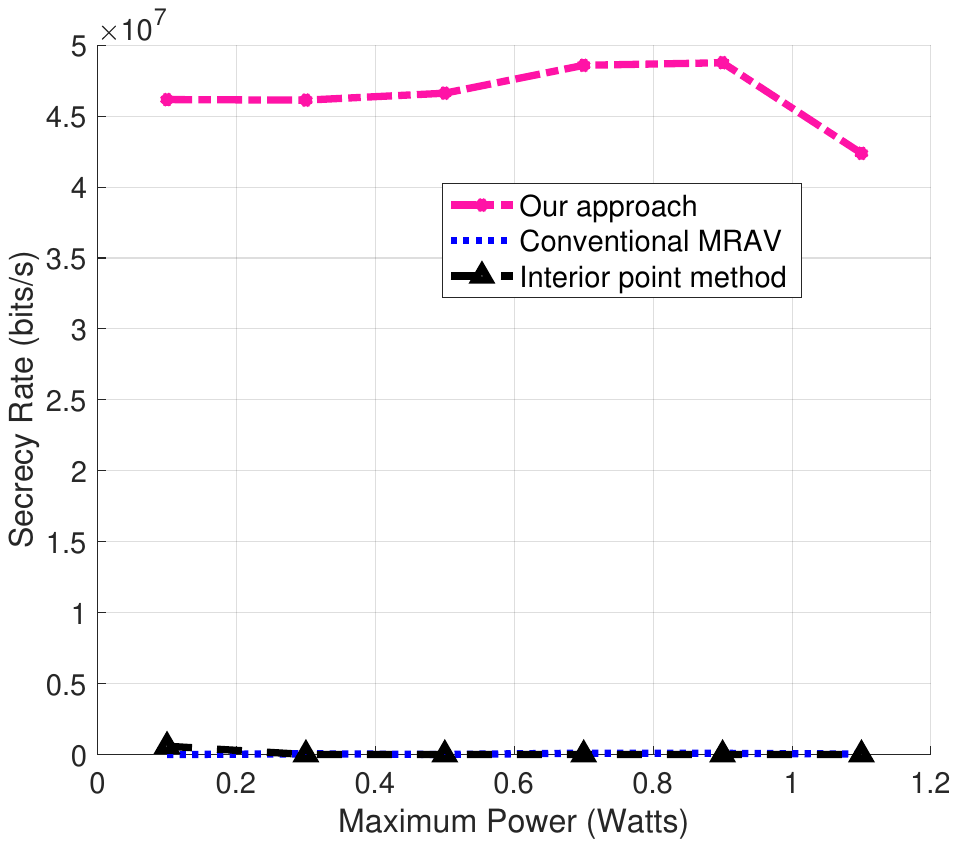}
         \vspace*{-2.2em}
         \caption{With uncertainty about the eavesdropper's position.}
         \label{fig:uncertaintyPosition}
    \end{subfigure}
    \vspace{-0.5em}
    \caption{Secrecy rate as a function of maximum power.}
    \label{figNew}
\end{figure*}


Figure \ref{fig3} illustrates the optimal 3D positions of \ac{MRAV}-I and \ac{MRAV}-J, along with lines $\mathcal{L}_I$ and $\mathcal{L}_J$, representing the direction of maximum gain of the antennas. 

As depicted in the figure, the maximum gain of the antenna of the \ac{MRAV}-I for all approaches is strategically directed towards the legitimate user to ensure efficient communication. Conversely, the orientation of the maximum gain of the antenna for the cooperative jammer, $\mathcal{L}_J$, is distinctly oriented away from the legitimate user and directed towards the eavesdroppers. It is also important to note that the $\mathcal{L}_I$ obtained by our proposed approach always intersects with the legitimate user, ensuring that the maximum gain of the antenna of the \ac{MRAV}-I is always directed to the legitimate user, contributing to the maximization of the secrecy rate. 

Figure \ref{fig:withouJittering} presents the secrecy rate plotted against the maximum power for both our proposed approach and the two benchmarks without any jittering on the \acp{MRAV} orientation and with perfect knowledge of the eavesdroppers position. As seen in the figure, our proposed approach achieves a significant performance compared to the interior-point method. This substantial gain arises from the non-convexity of the secrecy rate function, whereby the interior-point method may become trapped in a local optimum, significantly deviating from the optimal performance. Moreover, the figure shows that the conventional \ac{MRAV} approach exhibits the poorest performance due to its inability to optimize antenna orientation, resulting in very low performance. 


Figures \ref{fig:withJittering} and \ref{fig:uncertaintyPosition} illustrate the impact of jittering and uncertainty about the eavesdroppers position on the secrecy rate obtained by our approach and the benchmarks. In Figure \ref{fig:withJittering}, we examine the scenario where we optimize the expectation of the secrecy rate over jittering but test for a random jittering value. Whereas, in Figure \ref{fig:uncertaintyPosition}, we optimize the secrecy rate with an estimated value of the eavesdroppers position and test for a random value of their positions. As shown in the figure, our proposed approach outperforms the benchmarks for both scenarios, followed by  interior-point method. 



\section{Conclusions}
\label{sec:conclusions}

In this paper, we explored the application of teams of omnidirectional \acp{MRAV} to enhance the secrecy rate among legitimate nodes. We leveraged the unique capacity of omnidirectional \acp{MRAV} to independently control their position and orientation. Our investigation demonstrated that by utilizing this capability of omnidirectional \acp{MRAV} and integrating cooperative jamming techniques, we can substantially enhance the security of communications with legitimate nodes in environments containing multiple malicious eavesdroppers. To the best of authors' knowledge, this study represents the initial endeavor to employ such \acp{MRAV} in addressing these types of problems. However, numerous inquiries persist regarding the utilization of these \acp{MRAV} in other physical layer security scenarios. We contend that these inquiries present novel research avenues within the communications community. 


\balance
\bibliographystyle{IEEEtran}
\bibliography{bibliography}

\end{document}